\documentclass{article} 
\usepackage[preprint]{colm2026_conference}

\usepackage{microtype}
\usepackage{hyperref}
\usepackage{url}
\usepackage{booktabs}
\usepackage{xspace}

\usepackage{lineno}
\ifcolmsubmission
\linenumbers
\fi
\usepackage{array}
\usepackage{makecell}
\usepackage{graphicx}
\usepackage{wrapfig}
\usepackage{multirow}
\usepackage{siunitx}
\usepackage{caption}
\usepackage[utf8]{inputenc}     
\usepackage{microtype}
\usepackage[table]{xcolor}
\usepackage{colortbl}
\usepackage{verbatim}
\usepackage{float}

\usepackage[T1]{fontenc}   
\usepackage[utf8]{inputenc}
\DeclareUnicodeCharacter{2032}{'}
\usepackage{textcomp}      
\usepackage{listings}
\usepackage{epigraph}

\lstdefinestyle{evalstyle}{
    basicstyle=\ttfamily\footnotesize,
    breaklines=true,
    postbreak=,              
    showstringspaces=false,
    upquote=true,            
    columns=flexible,
    keepspaces=true,
    literate={—}{--}1 {“}{``}1 {”}{''}1 {’}{'}1, 
}

\definecolor{darkblue}{rgb}{0, 0, 0.5}
\hypersetup{colorlinks=true, citecolor=darkblue, linkcolor=darkblue, urlcolor=darkblue}

\title{SocialVeil: Probing Social Intelligence of Language Agents under Communication Barriers}

\author{
Keyang Xuan$^{1}$, Pengda Wang$^{2}$, Chongrui Ye$^{1}$, Haofei Yu$^{1}$, Tal August$^{1}$, Jiaxuan You$^{1}$ \\
$^1$Siebel School of Computing and Data Science, University of Illinois Urbana-Champaign \\
$^2$Department of Psychological Sciences, Rice University
}

\newcolumntype{P}[1]{>{\centering\arraybackslash}p{#1}}
\newcolumntype{L}[1]{>{\raggedright\arraybackslash}p{#1}}
\newcommand{\socialveil}{\textsc{SocialVeil}\xspace}
\newcommand{\sotopia}{\textsc{Sotopia}\xspace}

\begin{document}

\ifcolmsubmission
\linenumbers
\fi

\maketitle

\begin{abstract}
Large language models (LLMs) are increasingly evaluated in interactive environments to test their social intelligence. 
However, existing benchmarks often assume idealized communication between agents, limiting our ability to diagnose whether LLMs can maintain and repair interactions in more realistic, imperfect settings.
To close this gap, we present \socialveil\footnote{Our code and data are available at \url{https://github.com/ulab-uiuc/social-veil}.}, a social learning environment that can simulate social interaction under cognitive-difference-induced communication barriers. 
Grounded in a systematic literature review of communication challenges in human interaction, \socialveil introduces three representative types of such disruption, \emph{semantic vagueness}, \emph{sociocultural mismatch}, and \emph{emotional interference}. 
We also introduce two barrier-aware evaluation metrics, \emph{unresolved confusion} and \emph{mutual understanding}, to evaluate interaction quality under impaired communication. 
Experiments across 720 scenarios and four frontier LLMs show that barriers consistently impair performance, with mutual understanding reduced by over 45\% on average, and confusion elevated by nearly 50\%. 
Human evaluations validate the fidelity of these simulated barriers (ICC$\approx$0.78, Pearson r$\approx$0.80). 
We further demonstrate that adaptation strategies (Repair Instruction and Interactive learning) only have a modest effect far from barrier-free performance. 
This work takes a step toward bringing social interaction environments closer to real-world communication, opening opportunities for exploring the social intelligence of LLM agents.
\end{abstract}

\section{Introduction}
\vspace{-3mm}
\epigraph{\emph{``The production of meaning, rather than the production of messages.''}}{\citet{Barnlund1970Transactional}}
\vspace{-3mm}

As this quote demonstrates, communication is inherently dynamic and marked by uncertainty \citep{feldmanhall2019resolving}. 
Rather than simply exchanging facts, communication involves navigating ambiguity, managing relational dynamics, and repairing misunderstandings (e.g., \citealp{clark1991grounding}).
For example, if it is among close friends, a vague or implicit expression of an idea can often still be correctly interpreted based on shared contextual knowledge and prior experience; but among complete strangers, due to the lack of common background and mutual understanding, such vague expressions are more likely to lead to misunderstandings or communication failures (e.g., \citealp{grice1975logic}).

These breakdowns are not merely incidental—they often arise from deeper, systematic influences that shape how messages are sent and received \citep{clark1991grounding,schegloff1977preference}. 
We define these systematic factors that hinder mutual understanding in dialogue as \textbf{communication barriers} (e.g., \citealp{lunenburg2010communication}).
Recognizing their role in shaping agent behavior is essential for socially-aware AI systems operating in complex environments, as communication barriers expose subtle failure modes overlooked by aggregate metrics and provide diagnostic insight for robust and responsible deployment \citep{song2025survey}.

However, constructing a principled framework for simulating communication barriers is challenging due to several aspects: 
1) \textbf{Intractable Taxonomy.} 
Barriers manifest at many levels, from perceptual-level acoustic interference (e.g., \citealp{cherry1953some}) to discourse-level breakdowns (e.g., \citealp{schegloff1977preference}).
Existing research still lacks a well-structured and literature-supported taxonomy to reliably guide systematic investigation \citep{sap2019socialiqa}.
2) \textbf{Realism–Control Tradeoff.} 
Barriers must remain faithful to social practice yet also be instantiated in a controlled and reproducible manner \citep{aher2023using}.
Naive noise injection often breaks realism, whereas free-form prompts tend to sacrifice consistency and comparability.
3) \textbf{Metric Insufficiency.} 
The presence of a barrier does not always entail failure of social interaction; an agent may still accomplish its goal through brute-force strategies while undermining the relationship or mutual understanding \citep{brown1987politeness}. 
Thus, a barrier-aware environment must evaluate not only task success but also incorporate metrics that capture the broader effects of barriers.

\begin{figure*}[t]
  \centering
  \vspace{-6mm}
  \includegraphics[height=0.26\textheight]{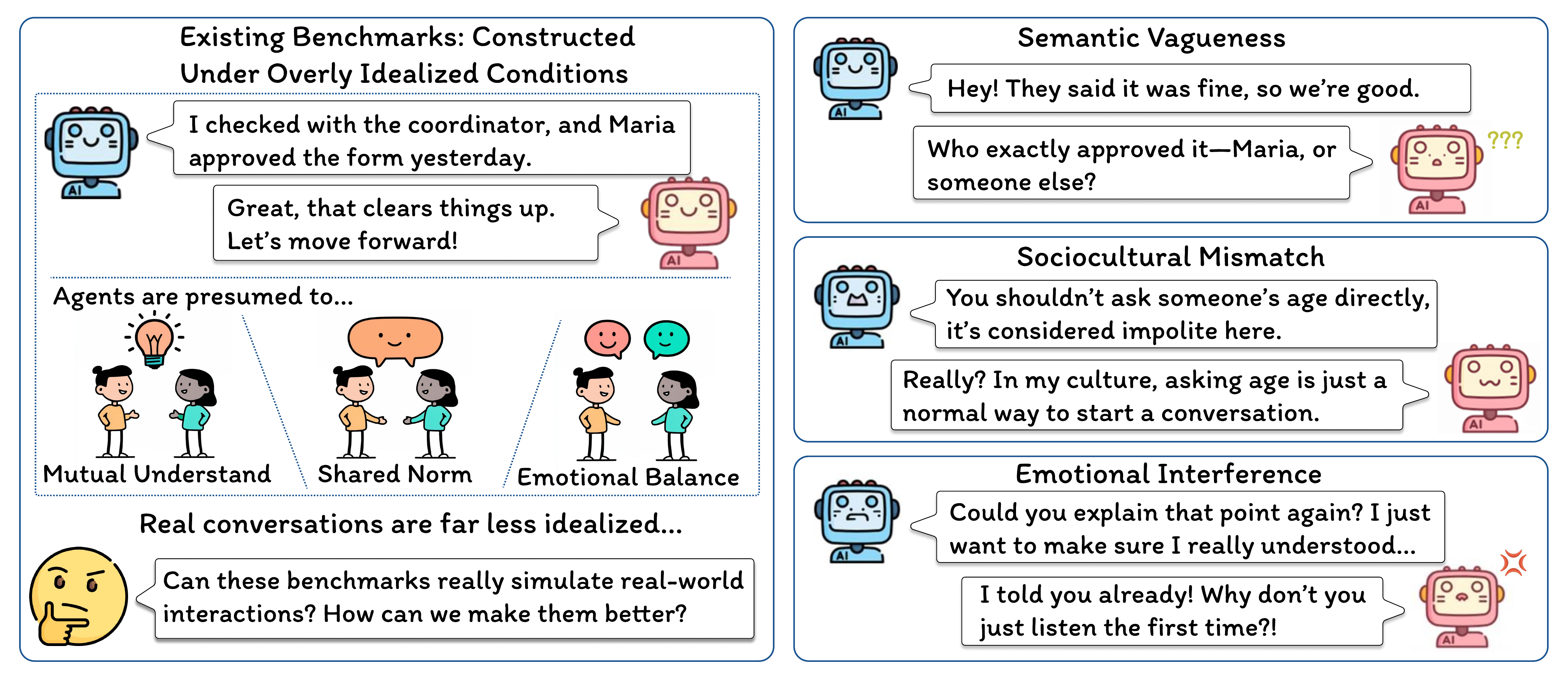}
    \caption{\textbf{Existing social benchmarks typically assume idealized conditions.} A comparison between existing benchmarks scenarios and real-world conversations, highlighting challenges in semantic vagueness, sociocultural mismatch and emotional interference.}
  \label{fig:motivation}
  \vspace{-4.5mm}
\end{figure*}

Recent studies have introduced interactive environments and benchmarks to assess agents' social intelligence capability (e.g., \citealp{chen2024socialbench,agentsense2024,zhou2023sotopia}). 
These works are typically constructed under highly idealized conditions.
Agents are presumed to share the same linguistic assumptions, sociocultural norms, and emotional registers, thereby overlooking the vagueness, misalignment, and disruptions that pervade real interaction.
As discussed previously and illustrated in Figure~\ref{fig:motivation}, actual conversational settings are far less idealized, where communication barriers often cause misunderstandings or conflicts, creating a more dynamic and complex interactional environment.

To emulate real-world settings, we introduce \socialveil, a framework for creating interactive social learning environments that evaluates agents' social intelligence under communication barriers. Through a systematic literature review, we identify three communication barriers rooted in cognitive factors \citep{sperber1986relevance}: \emph{Semantic Vagueness}, \emph{Sociocultural Mismatch}, and \emph{Emotional Interference}. 
We construct 180 episodes for each of the three barrier types as well as for a barrier-free baseline, resulting in a total of four sets of episodes. 
These scenarios are adapted from \sotopia \citep{zhou2023sotopia}.
To capture communication barriers' impact beyond social interaction task completion, we introduce a barrier-aware evaluation protocol and conduct a comprehensive evaluation of four frontier LLMs and verify their validity through comprehensive human evaluation. We further implement targeted interventions, such as repair-oriented instruction and the interactive learning framework, to enhance the agent's ability to engage in social interactions in barrier scenarios.

\textbf{Main Discoveries.}
Our analysis demonstrates that communication barriers consistently impair agents’ social intelligence capabilities.
For example, \emph{Semantic Vagueness} often prevents agents from establishing shared context, leading to substantial declines in mutual understanding (58\%$\downarrow$), and \emph{Emotional Interference} often disrupts relationship quality (49\%$\downarrow$).
We further validate the fidelity of simulated barriers and the reliability of our evaluation protocol through human evaluation:
annotators have demonstrated strong inter-rater reliability (\emph{avg ICC $\approx$ 0.78}), were able to accurately identify barrier types (\emph{avg Accuracy $\approx$ 68\%}), and showed strong alignment with automatic metrics (\emph{avg Pearson’s $r \approx$ 0.80}).
Furthermore, our exploration of potential adaptation strategies reveals that neither static nor dynamic interventions can effectively counteract these barriers, underscoring that \socialveil presents a fundamental challenge that transcends simple heuristic or supervised adjustments.

Overall, our contributions are as follows:
1) We introduce \socialveil, a barrier-aware, interactive social learning environment for simulating and evaluating LLM agents’ social intelligence under barrier scenarios.
2) We propose an automated, barrier-aware evaluation protocol that complements conventional goal-oriented measures by explicitly capturing whether agents can maintain interaction and repair misunderstandings under communication barriers.
3) We demonstrate that barriers simulated in \socialveil induce effects that align with their real-world counterparts, validating the framework’s fidelity as a proxy for studying real-world interaction.
4) We explore adaptation strategies for enhancing agents’ performance under barriers, showing that while repair instructions are largely ineffective, interactive learning yields steady yet limited improvements, highlighting both the promise of adaptive training and the remaining gap to human-level resilience.
\section{SocialVeil: A Barrier-Aware Social Learning Environment}
\label{sec:method}

\begin{figure*}[t]
  \centering
  \vspace{-6mm}
  \includegraphics[width=\textwidth]{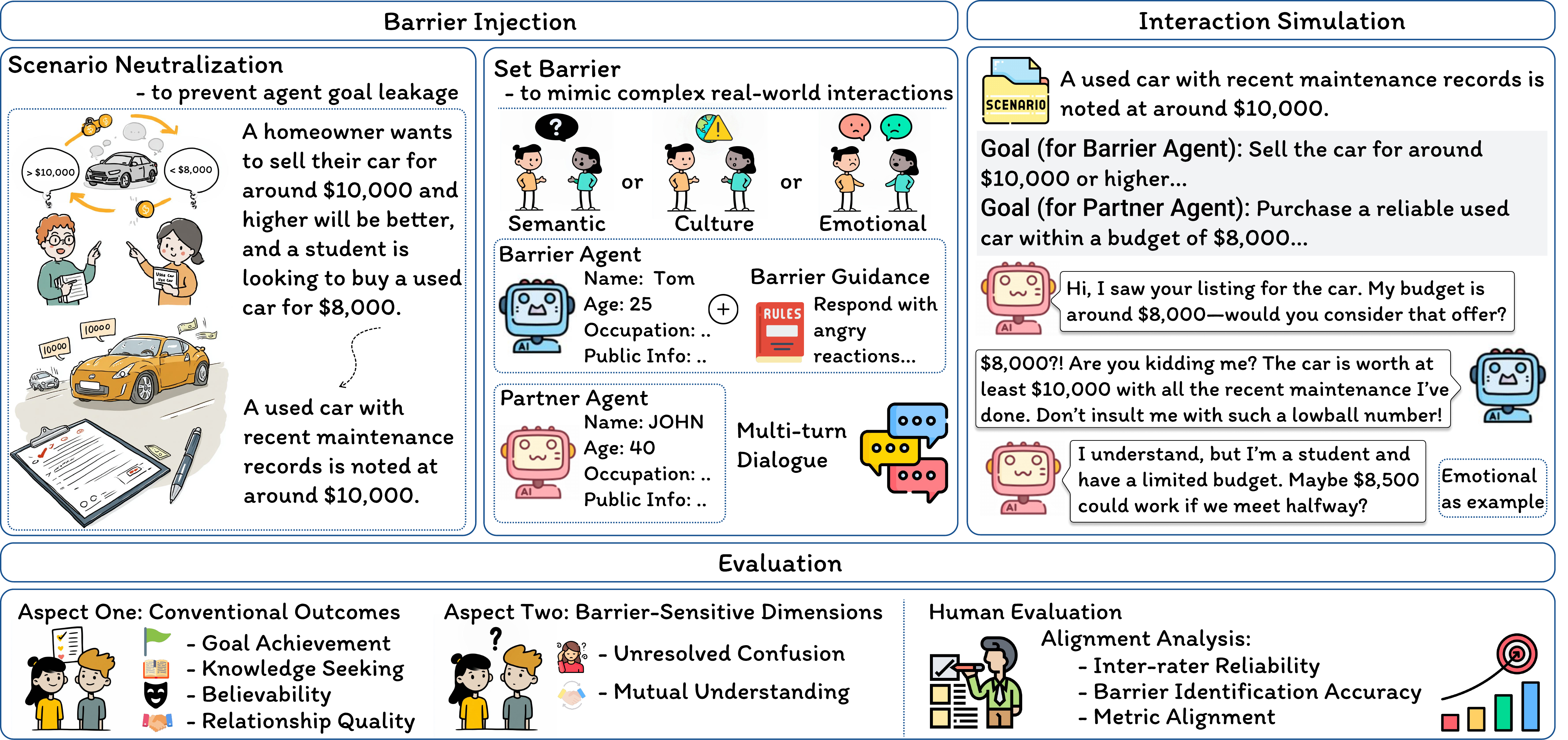}
  \caption{\textbf{Overview of \socialveil.} 
  The pipeline consists of three stages: 
  (\textbf{Barrier Injection}) scenarios are neutralized to remove bias and barriers are systematically injected; 
  (\textbf{Interaction Simulation}) agents engage in multi-turn dialogue under barrier conditions; 
  (\textbf{Evaluation}) agent performance is assessed using automatic metrics (goal achievement, knowledge seeking, believability, relationship quality, confusion, mutual understanding) and human evaluation (alignment analysis and barrier navigation success).}
  \vspace{-2mm}
  \label{fig:framework}
\end{figure*}

To better reflect real-world conditions, we propose a barrier-aware, socially interactive environment (\socialveil) with the following desiderata:
1) \textbf{Task Agnostic.} Barriers should preserve social goals and context.
2) \textbf{Structured Disruptions.} Barriers should be systematically designed to induce a specific level of disruption. 
3) \textbf{Barrier-aware Evaluation.} Environment must support evaluation beyond goal-oriented dimensions to capture the communication failures induced by the barriers.
An overview of \socialveil is shown in Figure~\ref{fig:framework}. 
We first introduce the creation of the barrier taxonomy (§\ref{par_taxonomy}), followed by their implementation (§\ref{par_barrierdesign}), simulation setup (§\ref{par_simulation}) and evaluation protocol (§\ref{par_eval}).

\begin{table}[ht]
\centering
\resizebox{\textwidth}{!}{%
\begin{tabular}{@{}P{2.2cm} L{6.0cm} L{3.5cm} L{5.7cm}@{}}
\toprule
\textbf{Barrier Type} & \textbf{Definition} & \textbf{Real-world Example} & \textbf{Theoretical Grounding} \\
\midrule
\textbf{\makecell[tc]{Semantic\\Vagueness}} &
Explicit referents are substituted with indeterminate pronouns or empty placeholders, leaving interpretation underspecified and prone to ambiguity. &
\textit{``It might work... you know what I mean.''} &
Pragmatics~\citep{grice1975logic}; 
Hedges~\citep{lakoff1973hedges};
Fuzzy logic~\citep{zadeh1965fuzzy} \\
\midrule
\textbf{\makecell[tc]{Sociocultural\\Mismatch}} &
Cultural differences in communication styles lead to misaligned interpretations and hinder explicit understanding. &
\textit{``We’ll think about it.'' Taken as postponement, but meant as refusal.} &
Politeness~\citep{brown1987politeness}; 
Context theory~\citep{hall1976beyond};  
Linguistic relativity~\citep{sapir1929status}\\
\midrule
\textbf{\makecell[tc]{Emotional\\Interference}} &
Affective intensity overrides informational clarity, displacing task-relevant content with expressive overflow. &
\textit{``I’m too upset to explain—just figure it out yourself!''} &
Attention~\citep{eysenck2007anxiety};
Emotion regulation~\citep{gross1998emerging}; Appraisal~\citep{lerner2000beyond};  \\
\bottomrule
\end{tabular}}
\caption{\textbf{Three types of communication barriers in \socialveil}, theoretically grounded and operationalized to reflect real-world interaction patterns.}
\label{tab:taxonomy}
\end{table}
\vspace{-1.0em}

\subsection{Barrier Taxonomy}  
\label{par_taxonomy}
Although external, physical barriers such as loud noise can obstruct the transmission of information, this study focuses on cognitive factors that hinder understanding, reasoning, and decision-making.
Through a systematic review of research on interaction and communication, we identified three major categories of communication barriers induced by cognitive factors: \textbf{\emph{Semantic Vagueness}}, ambiguity from vague pronouns or unspecified placeholders; \textbf{\emph{Sociocultural Mismatch}}, misaligned interpretations across cultural communication styles; and \textbf{\emph{Emotional Interference}}, affective intensity that obscures task content and negatively influences performance.
Table~\ref{tab:taxonomy} provides definitions, examples, and theoretical grounding for each barrier (details of the literature review presented in Appendix~\ref{sec: barrier}).

\subsection{Barrier Design}
\label{par_barrierdesign}
In \socialveil, we instantiate communication barriers unilaterally. Let $\mathcal{B}$ be the barrier set defined in §\ref{par_taxonomy};
One agent, designated as the barrier agent, communicates under a chosen barrier condition, while the partner agent remains in standard settings. 
Each barrier $b\in \mathcal{B}$ is instantiated by composing a style prompt $P_b$ with a parameterization $R_b$ over four operational dimensions: \textbf{Narrative Stance} (\emph{overall communication style, e.g., indirectness or emotion-focus}), \textbf{Interaction Tactics} (\emph{linguistic devices such as vague words, placeholder nouns, indirect refusals}), \textbf{Confusion Mechanisms} (\emph{behaviors that block understanding, e.g., withholding confirmation}), 
and \textbf{Exemplar Templates} (\emph{example patterns for reproducibility}).

In practice, each barrier $b$ is implemented through a two-layer design. The style prompt $P_b$ encodes a high-level directive (\emph{e.g., ``overuse pronouns and ellipses'' for semantic vagueness}), while the parameterization $R_b$ specifies quantitative cues that render the behavior reproducible. During simulation, $P_b$ and $R_b$ are utilized only for the barrier agent, whereas the partner agent remains unmodified ($P_b = \emptyset$, $R_b=\emptyset$), ensuring the barrier is the sole source of disruption and making evaluation controlled and repeatable.

\subsection{Simulation Setup}
\label{par_simulation}
\paragraph{Episode Design.}  
An \socialveil episode is a two-agent role-play in which each agent is assigned a private social goal and a role profile. 
One agent, designated as the barrier agent, communicates under a chosen barrier condition, while the other, the partner agent, remains unmodified. This asymmetric design reflects natural human scenarios: For instance, a colleague whose indirect style obscures intent, or a teammate whose emotions color their contributions. To construct episodes, we adapt scenarios from existing social benchmarks and apply a neutralization step to their public scenario descriptions, which may otherwise leak the agent's private goals. Specifically, we use GPT-4o~\citep{hurst2024gpt} to rewrite each scenario description following fixed instructions (Appendix~\ref{subsec:neu}) to eliminate goal-related hints. This ensures that both agents share the same public social context but cannot unintentionally infer each other’s private goals. Formally, an episode is defined as
\begin{equation}
E = (\mathcal{A}_b, \mathcal{A}_p, g_b, g_p, p_p, p_b, b),
\end{equation}
where $\mathcal{A}_b$ and $\mathcal{A}_p$ denote the barrier and partner agents, $g_b$ and $g_p$ their respective goals, $p_b$ and $p_p$ their role profiles, and $b$ the injected barrier type. 

\paragraph{Utterance Generation.}  
At each dialogue turn $t$, let $h_t$ denote the history of all utterances prior to $t$. Each agent $\mathcal{A}i$ generates an utterance $u_{t,i}$ conditioned on $h_t$, its goal $g_i$, and profile $p_i$. For the barrier agent $\mathcal{A}_b$, the instruction $I$ is augmented with the barrier specification $b$:
\begin{equation}
u_{t,b} \sim \pi_\theta(\cdot \mid h_t, g_b, p_b, I \oplus b),
\end{equation}
The partner agent $\mathcal{A}_p$, by contrast, generates from the unmodified instruction:
\begin{equation}
u_{t,p} \sim \pi_\theta(\cdot \mid h_t, g_p, p_p, I),
\end{equation}
This unilateral setup reflects natural conversational asymmetry and enables evaluation of the partner agent’s performance under impaired communication. Episodes are capped at 20 turns or conclude earlier if an agent chooses to exit.

\subsection{Evaluation Protocol}
\label{par_eval}
A barrier-aware social environment must judge both task success and social competence under impaired communication. Therefore, \socialveil evaluates social interaction along two complementary aspects:
The first aspect focuses on goal-oriented dimensions such as goal completion, relationship quality, and knowledge, which are widely measured in previous social interaction research~\citep{zhou2023sotopia}.
The second aspect introduces barrier-aware dimensions, which directly target the communicative disruptions induced by barriers.
Specifically, we introduce 1) \textbf{\emph{Unresolved Confusion}} to quantify the extent to which ambiguity remains at the end of the dialogue (five-point Likert scale, from incoherent to fully resolved), and 2) \textbf{\emph{Mutual Understanding}} to capture the degree of convergence on shared context and goals (five-point Likert scale, from complete misalignment to full alignment). The details of the evaluation protocol implementation are shown in Appendix~\ref{sec: Eval}.
\section{Research Questions and Experiment Setup}
\label{sec:RQ}
Our goal is to develop a systematic framework for simulating communication barriers and to leverage it as a diagnostic tool to investigate the social resilience of language agents.
Therefore, we frame our study in three guiding questions:
1) \textbf{Barrier Validity.} Do injected barriers reliably create structured disruptions?
2) \textbf{Barrier Effects.} How do barriers affect the performance of LLM agents in social interaction?
3) \textbf{Barrier Adaptation.} Can agents be improved to handle communication barriers?
To answer these questions, we build a testbed of 180 episodes for each barrier type (\emph{Semantic Vagueness}, \emph{Sociocultural Mismatch}, and \emph{Emotional Interference}), along with a baseline condition without barriers, following the procedure described in §\ref{sec:method}. All episodes are from \sotopia scenarios \citep{zhou2023sotopia}. We follow its difficulty tags and report two splits: \textbf{All} (standard + hard) and \textbf{Hard} (hard-only). We employ GPT-4o-mini as the base model for the barrier agent.

To investigate barrier effects, we evaluate four partner agents spanning both proprietary and open-weight model families: \texttt{GPT-4o-mini}~\citep{hurst2024gpt}, \texttt{Qwen2.5-7B-Instruct}, \texttt{Qwen3-4B-Instruct}~\citep{yang2025qwen2.5}, and \texttt{Mistral-8B-Instruct}~\citep{ministral8B-instruct-2410}. The temperature is set to 0.7 to encourage response diversity.
Performance is measured using the evaluation protocol described in §\ref{par_eval}, where we use GPT-4o as the backbone of the evaluator model, with its temperature set to 0 for stable, deterministic judgments. We further validate our barrier and evaluation consistency across alternative backbones in Appendix~\ref{sec:Robust}.

To investigate whether agents can be made more resilient to communication barriers, we implement two adaptation strategies:
1) \textbf{Repair Instruction.} We first examine a direct, instruction-based intervention. 
This approach enhances the partner agent's meta-prompt with explicit guidance designed to reduce misunderstandings (e.g., \emph{``Actively ask clarifying questions and paraphrase to confirm understanding.''}) 
2) \textbf{Interactive Learning.} We adapt an interactive learning framework (e.g., \citealp{wang2024sotopia}). 
The process begins with behavior cloning (BC), where expert trajectories are generated from interactions using GPT-4o as partner agents and filtered for success using our evaluation protocol. 
The partner agent is initialized by imitating these trajectories through training:
\begin{equation}
\mathcal{L}(\theta) = - \mathbb{E}_{(h_t, u_t^*) \sim \mathcal{D}} \left[ \log \pi_\theta(u_t^* \mid h_t) \right],
\end{equation}
where $\mathcal{D}$ is the set of high-quality demonstrations and $u_t^*$ is the expert utterance given history $h_t$.  
Then, we apply self-reinforcement (SR), the trained agent engages with the fixed barrier agent to produce new dialogues, from which high-quality trajectories are again filtered and added to training. This iterative process allows the agent to progressively distill strategies for navigating barriers.
\vspace{-2mm}

\section{Experiment Results}

\begin{table*}[t]
  \centering
  
  \newcommand{\ci}[1]{$^{\color{gray}{.#1}}$}
  \newcommand{\graynum}[1]{\textcolor{gray}{#1}}
  \resizebox{\textwidth}{!}{
    \setlength{\tabcolsep}{2pt}
    \begin{tabular}{@{} l l *{12}{c} @{}}
      \toprule
      & & \multicolumn{6}{c}{\textbf{Sotopia-All}} & \multicolumn{6}{c}{\textbf{Sotopia-Hard}} \\
      \cmidrule(lr){3-8} \cmidrule(lr){9-14}
      \textbf{Model} & \textbf{Type} & BEL$\uparrow$ & REL$\uparrow$ & KNO$\uparrow$ & GOAL$\uparrow$ & Conf$\uparrow$ & Mutu$\uparrow$ & BEL$\uparrow$ & REL$\uparrow$ & KNO$\uparrow$ & GOAL$\uparrow$ & Conf$\uparrow$ & Mutu$\uparrow$ \\
      \midrule

      \rowcolor[gray]{.95} \textbf{GPT-4o-m} 
      & \graynum{Base} & \graynum{8.78}\ci{11} & \graynum{3.41}\ci{21} & \graynum{3.94}\ci{18} & \graynum{7.60}\ci{22} & \graynum{4.08}\ci{15} & \graynum{4.56}\ci{10} & \graynum{8.82}\ci{12} & \graynum{2.53}\ci{15} & \graynum{2.78}\ci{11} & \graynum{6.75}\ci{21} & \graynum{3.75}\ci{14} & \graynum{4.46}\ci{11} \\
      & Sem  & 7.58\ci{15} & 1.91\ci{14} & 2.89\ci{20} & 5.61\ci{25} & 1.48\ci{12} & 1.76\ci{14} & 7.32\ci{11} & 1.61\ci{13} & 2.71\ci{14} & 5.46\ci{20} & 1.57\ci{12} & 1.78\ci{11} \\
      & Soc  & 7.70\ci{12} & 1.95\ci{18} & 2.92\ci{15} & 5.35\ci{28} & 1.78\ci{11} & 2.41\ci{13} & 7.61\ci{10} & 1.64\ci{15} & 2.96\ci{16} & 5.32\ci{24} & 1.75\ci{11} & 2.28\ci{10} \\
      & Emo  & 7.76\ci{10} & 1.61\ci{11} & 2.97\ci{14} & 5.25\ci{20} & 1.51\ci{10} & 2.03\ci{11} & 7.42\ci{12} & 1.03\ci{10} & 2.93\ci{15} & 4.96\ci{21} & 1.46\ci{10} & 1.89\ci{12} \\
      \addlinespace[0.2em]
      
      \rowcolor[gray]{.95} \textbf{Qwen2.5-7b} 
      & \graynum{Base} & \graynum{8.48}\ci{12} & \graynum{3.17}\ci{21} & \graynum{3.79}\ci{25} & \graynum{7.48}\ci{26} & \graynum{4.06}\ci{14} & \graynum{4.45}\ci{12} & \graynum{8.33}\ci{14} & \graynum{1.72}\ci{12} & \graynum{2.73}\ci{11} & \graynum{5.68}\ci{20} & \graynum{3.16}\ci{12} & \graynum{3.89}\ci{11} \\
      & Sem  & 7.37\ci{13} & 1.91\ci{14} & 2.63\ci{19} & 5.99\ci{23} & 1.61\ci{11} & 1.91\ci{15} & 6.96\ci{15} & 1.46\ci{11} & 2.25\ci{10} & 5.25\ci{18} & 1.21\ci{10} & 1.39\ci{12} \\
      & Soc  & 7.65\ci{13} & 2.04\ci{22} & 2.79\ci{17} & 5.71\ci{29} & 1.85\ci{13} & 2.45\ci{17} & 7.28\ci{12} & 1.64\ci{14} & 2.61\ci{13} & 5.17\ci{21} & 1.57\ci{11} & 1.93\ci{13} \\
      & Emo  & 7.56\ci{08} & 1.57\ci{14} & 2.64\ci{16} & 5.47\ci{19} & 1.63\ci{08} & 2.16\ci{13} & 7.28\ci{10} & 0.89\ci{11} & 2.28\ci{12} & 4.75\ci{19} & 1.39\ci{10} & 1.78\ci{11} \\
      \addlinespace[0.2em]

      \rowcolor[gray]{.95} \textbf{Qwen3-4b} 
      & \graynum{Base} & \graynum{8.64}\ci{13} & \graynum{3.12}\ci{22} & \graynum{3.88}\ci{24} & \graynum{7.73}\ci{24} & \graynum{3.72}\ci{16} & \graynum{4.30}\ci{11} & \graynum{8.40}\ci{12} & \graynum{1.96}\ci{11} & \graynum{2.93}\ci{15} & \graynum{6.57}\ci{20} & \graynum{3.14}\ci{11} & \graynum{4.03}\ci{10} \\
      & Sem  & 7.80\ci{09} & 2.03\ci{16} & 2.85\ci{19} & 6.81\ci{27} & 1.97\ci{13} & 2.42\ci{16} & 7.61\ci{11} & 1.32\ci{13} & 2.28\ci{11} & 6.04\ci{22} & 1.68\ci{10} & 1.86\ci{12} \\
      & Soc  & 7.89\ci{10} & 2.04\ci{15} & 3.10\ci{18} & 6.45\ci{24} & 2.26\ci{14} & 3.02\ci{16} & 7.64\ci{10} & 1.28\ci{12} & 3.03\ci{14} & 5.96\ci{21} & 2.03\ci{11} & 2.64\ci{11} \\
      & Emo  & 7.94\ci{08} & 1.72\ci{14} & 3.06\ci{14} & 6.48\ci{28} & 1.94\ci{07} & 2.64\ci{11} & 7.75\ci{08} & 1.21\ci{11} & 2.75\ci{11} & 5.78\ci{23} & 1.67\ci{09} & 2.32\ci{10} \\
      \addlinespace[0.2em]

      \rowcolor[gray]{.95} \textbf{Mistral-8b} 
      & \graynum{Base} & \graynum{7.73}\ci{15} & \graynum{2.84}\ci{20} & \graynum{3.74}\ci{22} & \graynum{6.83}\ci{21} & \graynum{3.23}\ci{18} & \graynum{3.54}\ci{14} & \graynum{7.73}\ci{16} & \graynum{2.42}\ci{14} & \graynum{3.39}\ci{18} & \graynum{6.22}\ci{21} & \graynum{2.61}\ci{15} & \graynum{3.25}\ci{14} \\
      & Sem  & 7.01\ci{16} & 1.56\ci{11} & 2.41\ci{15} & 3.91\ci{24} & 1.07\ci{10} & 1.13\ci{12} & 6.81\ci{15} & 1.32\ci{11} & 2.25\ci{13} & 3.50\ci{19} & 1.01\ci{08} & 1.00\ci{10} \\
      & Soc  & 7.81\ci{14} & 2.28\ci{19} & 3.02\ci{17} & 5.31\ci{25} & 1.52\ci{12} & 1.85\ci{14} & 6.45\ci{12} & 2.07\ci{16} & 3.07\ci{15} & 5.07\ci{22} & 1.21\ci{11} & 1.43\ci{11} \\
      & Emo  & 7.47\ci{12} & 1.39\ci{10} & 2.69\ci{14} & 4.48\ci{21} & 1.26\ci{10} & 1.42\ci{11} & 6.48\ci{09} & 1.28\ci{11} & 2.79\ci{11} & 4.03\ci{14} & 1.11\ci{09} & 1.18\ci{10} \\
      \bottomrule
    \end{tabular}
  }
  \caption{\textbf{Barriers consistently degrade agent performance compared to the baseline across all models.} We observe: 1) Semantic vagueness most severely impairs mutual understanding, 2) Emotional interference disproportionately harms relationships, and 3) Sociocultural mismatch induces persistent confusion. For information on the metrics, see Appendix~\ref{sec: Eval}.}
  \label{tab:main_result}
  \vspace{-4mm}
\end{table*}

\subsection{Are the Created Barrier Valid?}
\label{sec: RQ1}
In sociolinguistics, barriers are viewed as structured phenomena that systematically distort interactional signals~\citep{clark1991grounding, tannen2005conversational}. Thus, a valid simulated barrier should induce structured distribution shifts in communication representations.

\begin{wrapfigure}[14]{r}{0.35\columnwidth}
  \centering
  \vspace{-15pt}
  \includegraphics[width=1.0\linewidth,trim=6 5 6 6,clip]{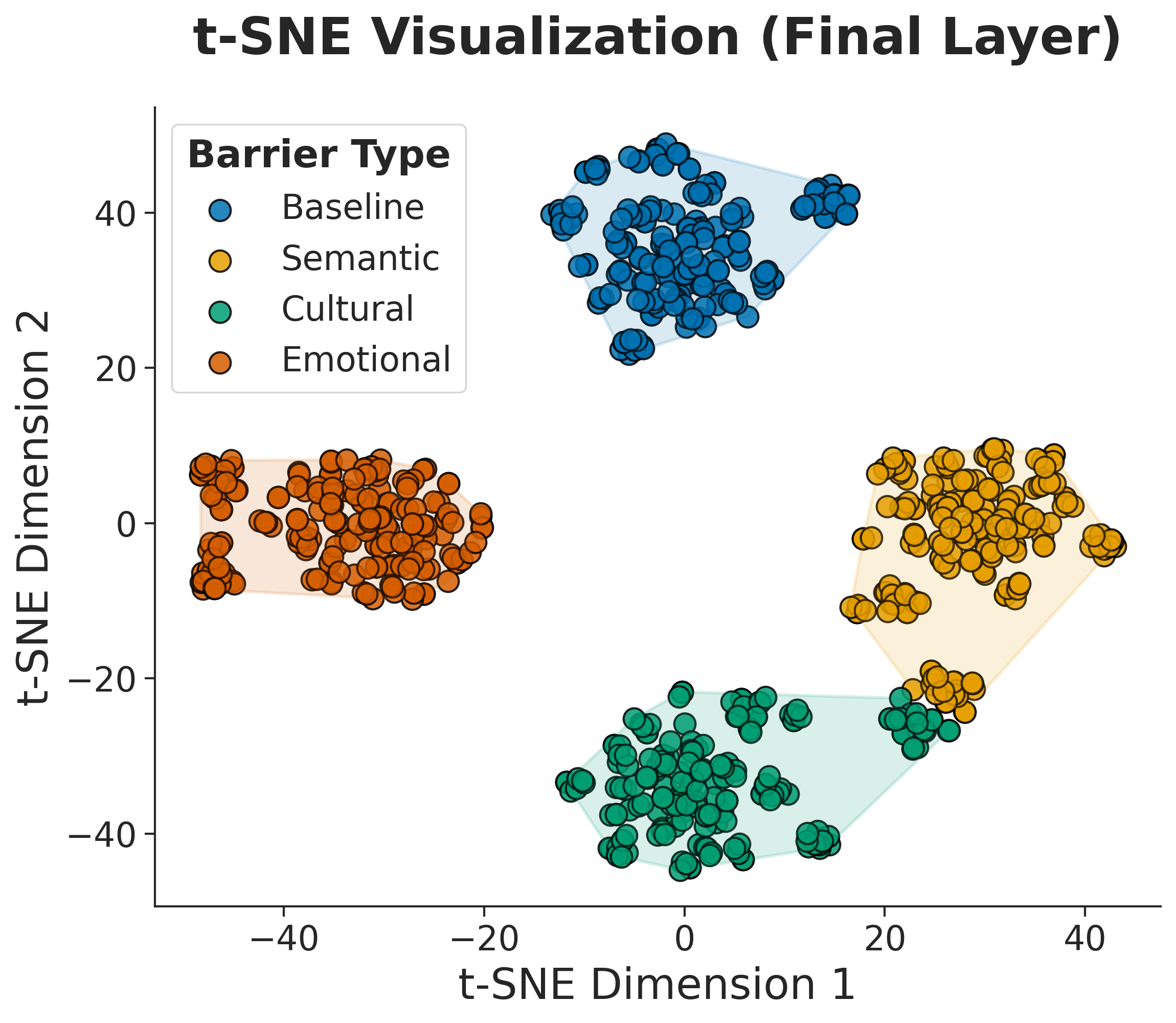}
  \vspace{-10pt} 
  \caption{\small \textbf{t-SNE visualization of the final-layer representations from Qwen2.5-7B-Instruct.}}
  \label{fig:pca-b}
\end{wrapfigure}

We tested this by probing the hidden states of \texttt{Qwen2.5-7B-Instruct} in the baseline and three barrier conditions. 
As shown in Figure~\ref{fig:pca-b}, barrier conditions form distinct and compact clusters in the t-SNE space, with a clear separation from baseline points. 
This indicates that barriers are encoded in the model’s internal representations as structured modes of variation, rather than as random noise. Moreover, the three different barrier types also form clearly separated clusters with no overlap, further validating the design rationality of the simulated barriers. 
This mutual separation suggests that each barrier introduces a unique, consistent shift in communicative representation, reflecting distinct interactional distortions.

\vspace{-2mm}
\subsection{How do Barriers Affect Social Interaction?}
\label{sec: RQ2}
Table~\ref{tab:main_result} summarizes each model's performance across both baseline and barrier conditions. 
From these comprehensive results, we identify the following three key findings:

1) \textbf{Barriers consistently impair social interaction performance.} Across all evaluated models and multiple evaluation dimensions, the presence of barriers leads to significant and consistent performance degradation compared to the baseline. This detrimental effect holds across metrics from both goal-oriented dimensions and barrier-aware dimensions.

2) \textbf{Barrier types exhibit distinct patterns.} Each barrier produces a characteristic pattern of degradation. Semantic vagueness most severely disrupts mutual understanding (\emph{avg $-58\%$}), often preventing agents from converging on shared context. Emotional interference disproportionately damages the quality of the relationship (\emph{avg $-49\%$}), while sociocultural mismatch induces persistent confusion (\emph{avg $-49\%$}) with relatively mild relational effects. 

3) \textbf{Social reasoning is more fragile than goal pursuit.} Compared to goal completion and knowledge acquisition, which decline moderately under barriers ($20-30\%$), social dimensions suffer substantially greater degradation, with relationship quality dropping by an average of $45\%$ and mutual understanding declining by $52\%$ across all barrier types, which infers barriers primarily disrupt the subtle social reasoning required.

\renewcommand{\thefootnote}{\arabic{footnote}}
\setcounter{footnote}{0}

\vspace{-1mm}
\subsection{Can Agents Adapt to Communication Barriers?}
\label{sec: RQ3}
Table~\ref{tab:train_result} reports the comparison of agent social interaction performance under communication barriers between the default setting and the two adaptation strategies. We can draw three key findings: 

1) \textbf{Repair Instruction yields trivial performance improvements.} This result highlights two points. First, overcoming communication barriers is not a trivial skill that can be invoked by an instruction-level guidance; it requires agents to detect when breakdowns occur, attribute them to a specific distortion, and deploy targeted repair strategies. Second, the failure of this strategy reflects the limitation of static, barrier-agnostic prompting: agents often resort to shallow repetitions or generic clarifications that do not resolve the underlying disruption. 

2) \textbf{Interactive Learning (BC+SR) yields consistent but limited improvements.} In contrast, interactive learning produces steady gains across all barrier types and relieves agents' struggle with social interaction in barrier cases. Yet the improvements remain modest (\emph{avg 10-20\%}), such that the enhancement over Repair Instruction is incremental rather than dramatic, and performance still falls significantly short of the barrier-free baseline. 

3) \textbf{Both strategies show minimal impact on goal completion.} Notably, neither approach improves the GOAL scores relative to the baseline, suggesting that adaptation strategies may guide agents to focus primarily on mitigating barriers rather than social objectives. This indicates a potential trade-off where barrier-handling mechanisms divert cognitive resources away from goal-oriented behavior, highlighting the challenge of simultaneously maintaining task performance while addressing communication disruptions.

\begin{table}[t]
  \centering
  \captionsetup{labelsep=period, justification=raggedright, singlelinecheck=false}
  
  \begingroup
  \setlength{\tabcolsep}{2.2pt}
  \renewcommand{\arraystretch}{1.1}
  \footnotesize
  
  \newcommand{\graycell}[1]{\textcolor{gray}{#1}}

  \begin{tabular}{@{}ll ccc ccc ccc@{}}
    \toprule
    \multirow{2}{*}{\textbf{Model}} & \multirow{2}{*}{\textbf{Barrier}} &
      \multicolumn{3}{c}{\textbf{GOAL}} &
      \multicolumn{3}{c}{\textbf{Mutu}} &
      \multicolumn{3}{c@{}}{\textbf{Conf}} \\  
    \cmidrule(lr){3-5}\cmidrule(lr){6-8}\cmidrule(l){9-11} 
    & & {\graycell{Base}} & {Rep} & {BC+SR} & {\graycell{Base}} & {Rep} & {BC+SR} & {\graycell{Base}} & {Rep} & {BC+SR} \\  
    \midrule
    \multirow{3}{*}{Q2.5-7B}
      & Semantic       & \graycell{5.99} & 6.07 & 6.02 & \graycell{1.91} & 1.90 & \textbf{2.15} & \graycell{1.61} & 1.66 & \textbf{1.84} \\
      & Sociocult.     & \graycell{5.71} & 5.87 & 5.79 & \graycell{2.45} & 2.50 & \textbf{2.60} & \graycell{1.85} & 1.88 & \textbf{2.16} \\
      & Emotional      & \graycell{5.47} & 5.74 & \textbf{5.86} & \graycell{2.16} & 2.28 & \textbf{2.34} & \graycell{1.63} & 1.69 & \textbf{1.85} \\
    \midrule
    \multirow{3}{*}{Q3-4B}
      & Semantic       & \graycell{6.81} & 7.08 & \textbf{7.13} & \graycell{1.97} & 1.98 & \textbf{2.10} & \graycell{2.42} & 2.48 & \textbf{2.49} \\
      & Sociocult.     & \graycell{6.45} & 6.85 & 6.84 & \graycell{2.26} & 2.43 & \textbf{2.49} & \graycell{3.02} & 3.08 & \textbf{3.21} \\
      & Emotional      & \graycell{6.48} & 6.55 & 6.52 & \graycell{1.94} & 1.92 & \textbf{2.09} & \graycell{2.64} & 2.39 & 2.46 \\
    \bottomrule
  \end{tabular}
  \caption{\textbf{Effectiveness of Mitigation Strategies.} While instruction-level repair yields negligible improvements, interactive learning (BC+SR) offers consistent but modest gains. However, neither strategy fully restores agents to baseline proficiency.}
  \label{tab:train_result}
  \endgroup
\vspace{-1.5em}
\end{table}
\vspace{-2mm}
\section{Discussion}
\subsection{Analysis: Behavioral Alignment of Simulated Barriers}
To further analyze the effects of injected barriers, we move beyond quantitative metrics to conduct a behavioral analysis of the conversation trajectory.Inspired by previous work on evaluating the alignment of simulated behaviors~\citep{park2023generative, han2025personality, zou2024can}, we design a series of in-depth qualitative analyses to study the alignment between our simulated barriers and their real-world counterparts.

\textbf{Linguistic signatures.}
We first test whether barriers trigger systematic linguistic shifts consistent with real-world communication breakdowns. Empirically, we extracted four linguistic features from conversations: reference pronouns (e.g., ``it,'' ``that'')~\citep{Ariel1990AccessingNA}, hedging words (e.g., ``maybe,'' ``could'')~\citep{lakoff1973hedges}, sentiment polarity, and self-focus pronouns (e.g., ``I,'' ``my'')~\citep{pennebaker2011secret} and correlated them with evaluation protocol.

\begin{wrapfigure}[18]{r}{0.55\textwidth}
  \vspace{-1.0\baselineskip}
  \centering
  \includegraphics[width=\linewidth]{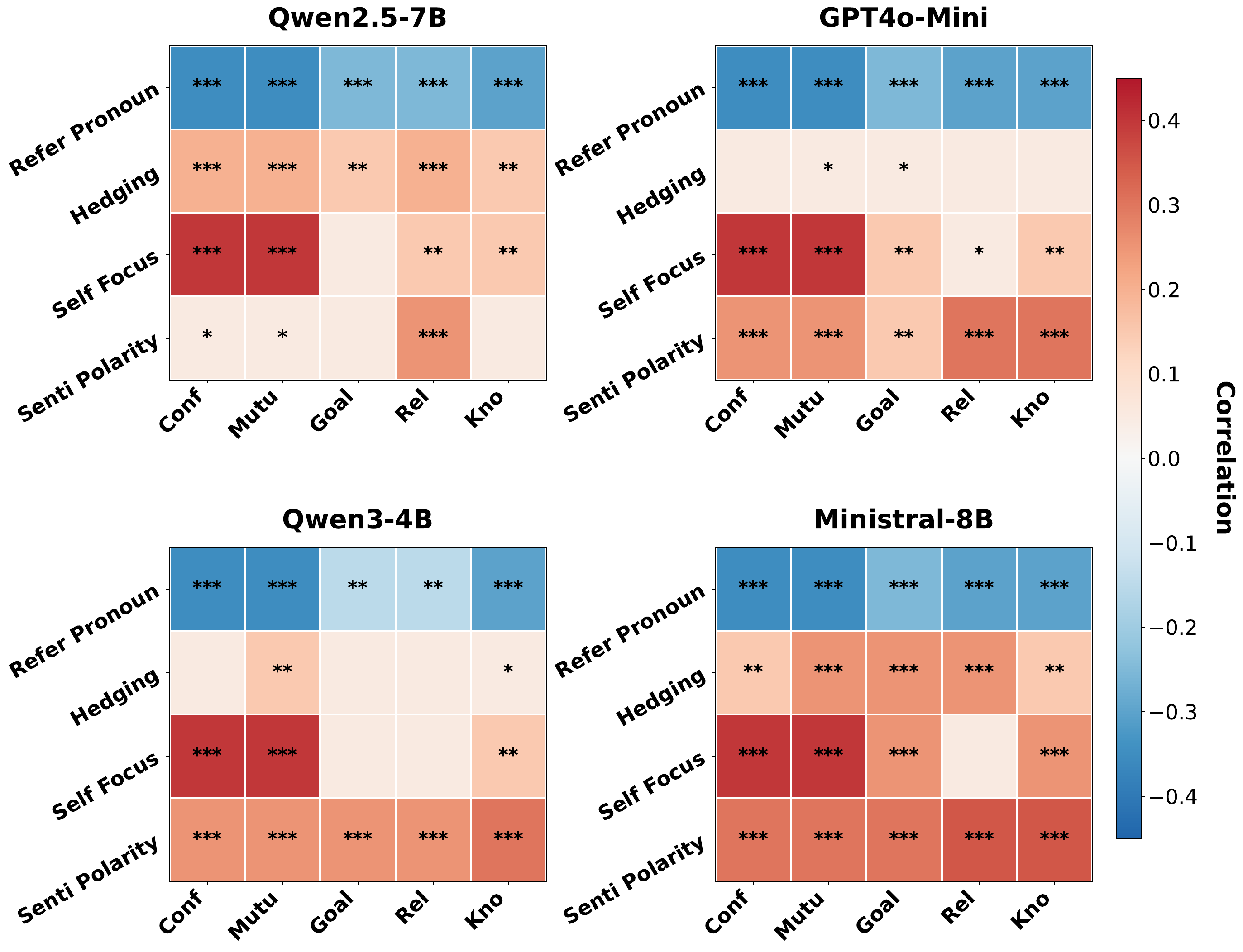}
  \caption{Linguistic signal and metrics correlations across four models. *$p$ < .05, **$p$ < .01, ***$p$ < .001.}
  \label{fig:corr}
  \vspace{-1.2\baselineskip}
\end{wrapfigure}

Figure~\ref{fig:corr} reveals two patterns: 1) Reference pronouns and self-focus are negatively associated with conversational quality, where they show correlation with higher confusion and lower mutual understanding; and 2) Sentiment polarity serves as a positive predictor of smooth interaction, where a more positive tone aligns with better relationships and greater goal attainment.

\begin{wrapfigure}[16]{l}{0.55\textwidth} 
  \centering
  \vspace{-1.2\baselineskip}
  \includegraphics[width=\linewidth]{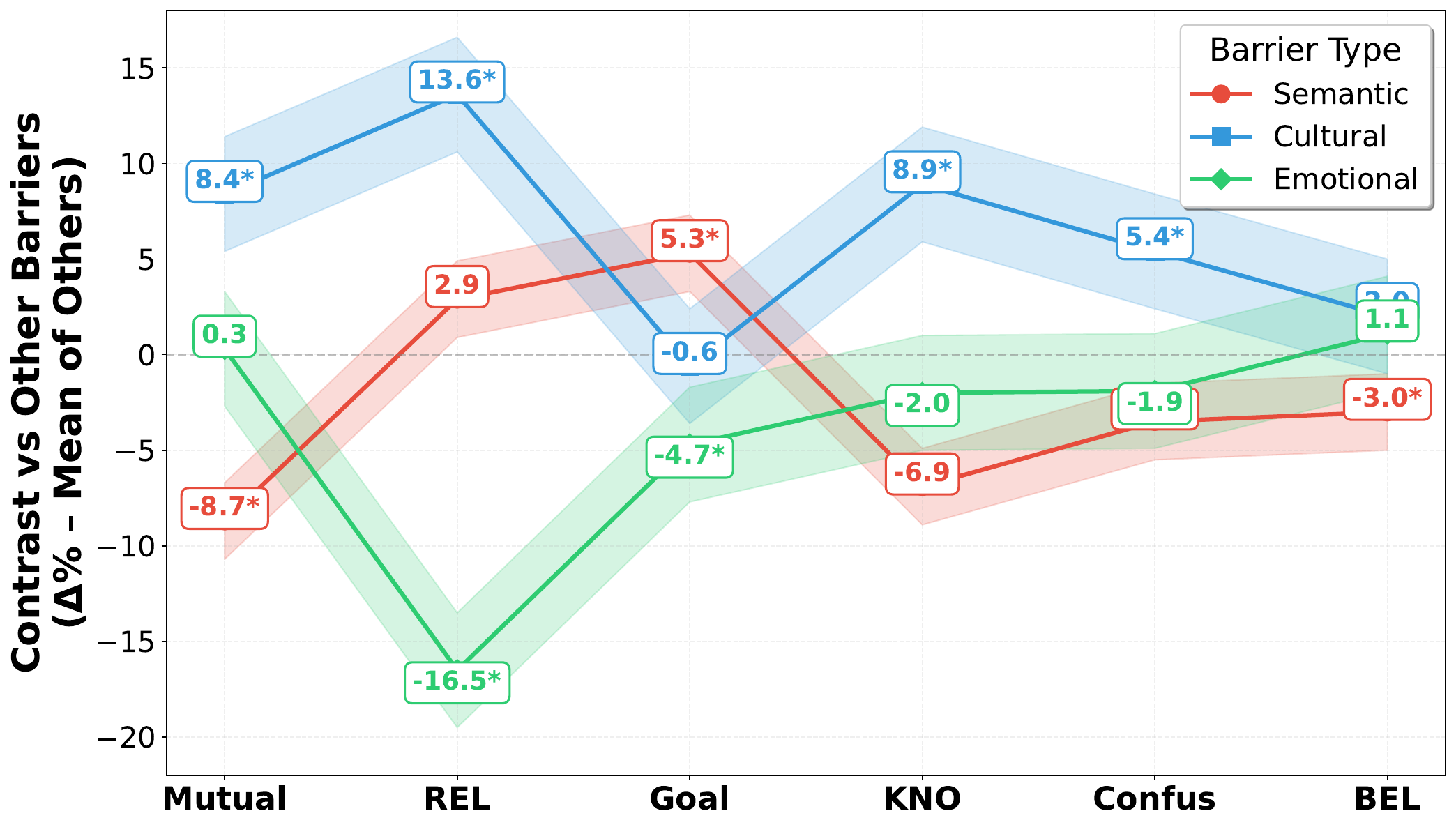}
  \caption{\textbf{Signatures of simulated communication barriers.} Data are shown as percentage deviations relative to other types *$p$ < .05.}
  \label{fig:barrier-signatures}
  \vspace{-1.0\baselineskip}
\end{wrapfigure}

\textbf{Barrier-specific effects.}
We quantify each barrier's unique effect as its deviation from the mean of the other two barriers on the same metric, model, and scenario, with bootstrap 95\% confidence intervals (Figure~\ref{fig:barrier-signatures}). The results further support that the effect of our simulated barriers aligns with their real-world counterparts: For example, semantic barriers most strongly impair Mutual Understanding, while emotional barriers disproportionately erode Relationship Quality, and Cultural barriers uniquely elevate Unresolved Confusion. These effects are consistent across models and statistically significant in the expected directions, demonstrating that the simulated barriers not only alter surface linguistic features but also reproduce distinct interactional disruptions characteristic of their real-world analogs.

\vspace{-3mm}

\subsection{Analysis: Human Evaluation}
To complement our automated metrics, we conduct a human evaluation to validate the model's ratings. 
Detailed procedures for human evaluation are provided in Appendix~\ref{sec: humaneval}. 
We focus on three key aspects: 1) \textbf{Inter-rater Reliability.} Assessing the consistency and reliability of human annotators; 2) \textbf{Barrier Identification Accuracy.} Evaluating whether human annotators can correctly identify the injected barriers; and 3) \textbf{Metric Alignment.} Examining how well the model-rated results correspond with human judgments.

\textbf{Inter-rater reliability.}
We calculated Fleiss’s Kappa for human annotation of barrier types, yielding a score of 0.38. 
Based on the classic interpretive framework of \citet{landis1977measurement}, this reflects fair agreement, bordering on moderate. 
Such values are common in complex, subjective multi-class annotation tasks. 
Prior studies report similar $\kappa$ ranges: \citet{castro-etal-2019-towards} reported values ranging from 0.23 to 0.59 for sarcasm detection, and \citet{callejas2008use} reported 0.32–0.42 for non-acted emotion identification. 
Thus, our result aligns well with similar findings in the existing literature, further supporting the overall reliability and consistency of our manual annotation process.

For rating metrics, each scenario was rated by three of six annotators. 
We used a one-way random effects model with single-measure ICC(1,k) to assess inter-rater reliability. 
\emph{Unresolved Confusion} showed ICC = 0.77, 95\% CI [0.68, 0.83], $F$(119, 240) = 4.26. 
\emph{Mutual Understanding} showed ICC = 0.79, 95\% CI [0.72, 0.85], $F$(119, 240) = 4.80.
Per \citet{cicchetti1994guidelines}, these values fall in the good range, suggesting reliable inter-rater consistency ($p$ $<$ .001).

\textbf{Barrier identification accuracy.}
We further reported the accuracy of human annotators in identifying barriers. 
Given that the sample contained only 120 scenarios, we employed cluster bootstrap resampling with replacement at the scenario level, repeated 1000 times, to evaluate the stability of the estimates and to construct 95\% percentile-based confidence intervals (95\% CI). 
From Figure~\ref{fig:human_type_acc}, the results showed that the overall accuracy of human annotators in barrier identification was 0.68 (95\% CI [0.63, 0.73]);
with accuracy 0.76 (95\% CI [0.67, 0.86]) for baseline barriers, 0.65 (95\% CI [0.54, 0.76]) for semantic barriers, 0.63 (95\% CI [0.53, 0.73]) for cultural barriers, and 0.67 (95\% CI [0.54, 0.78]) for emotional barriers. 
These results demonstrate that, across all barrier types, annotator accuracy was significantly above the level expected from random guessing, as all lower confidence interval bounds exceeded the binary chance baseline of 0.50.

\begin{table*}[htbp]
\centering
\setlength{\tabcolsep}{4pt}
\renewcommand{\arraystretch}{1.1}
\footnotesize
\begin{tabular}{@{}l*{8}{c}@{}}
\toprule
\multirow{2}{*}{\textbf{Barrier}} & \multicolumn{4}{c}{\textbf{Human Evaluation}} & \multicolumn{4}{c}{\textbf{Automated Evaluation}} \\
\cmidrule(lr){2-5} \cmidrule(lr){6-9}
& \multicolumn{2}{c}{CONFUSION} & \multicolumn{2}{c}{MUTUAL} & \multicolumn{2}{c}{CONFUSION} & \multicolumn{2}{c}{MUTUAL} \\
& Mean & 95\% CI & Mean & 95\% CI & Mean & 95\% CI & Mean & 95\% CI \\
\midrule
Baseline & 3.74 & [3.48, 4.00] & 3.84 & [3.62, 4.07] & 3.94 & [3.65, 4.24] & 4.47 & [4.27, 4.67] \\
Semantic & 1.45 & [1.30, 1.62] & 1.71 & [1.56, 1.87] & 1.54 & [1.32, 1.79] & 1.81 & [1.52, 2.13] \\
Sociocultural & 1.94 & [1.67, 2.23] & 2.25 & [2.01, 2.51] & 1.67 & [1.50, 1.82] & 2.24 & [1.96, 2.50] \\
Emotional & 1.70 & [1.53, 1.86] & 1.96 & [1.80, 2.11] & 1.67 & [1.50, 1.84] & 2.20 & [1.92, 2.50] \\
\bottomrule
\end{tabular}
\caption{\textbf{Mean ratings by barrier type with 95\% confidence intervals}. Our results show strong alignment between human annotators and automated evaluation in rating \emph{Mutual Understanding} and \emph{Unresolved Confusion}. Pearson correlation: Confusion $r{=}0.80$ (95\% CI [0.72, 0.86]), Mutual $r{=}0.79$ (95\% CI [0.71, 0.85]). $n{=}120$.}
\label{tab:human_auto_comparison_with_ci}
\end{table*}

\textbf{Metric alignment.}
We further compared the mean values and 95\% confidence intervals of \emph{Unresolved Confusion} and \emph{Mutual Understanding} between human annotators and the model under different barrier conditions. 
As shown in Table~\ref{tab:human_auto_comparison_with_ci}, the model’s performance was close to the average human ratings, with only small differences in the confidence intervals, indicating a high degree of consistency. 
In addition, the analysis of overall convergent validity revealed statistically strong correlations between human and model scores, with \emph{Unresolved Confusion} yielding a convergent validity of 0.80 (95\% CI [0.72, 0.86]) and \emph{Mutual Understanding} yielding a convergent validity of 0.79 (95\% CI [0.71, 0.85]). 
These results offer further support for the model’s reliability in evaluating the barrier-sensitive dimensions.
\section{Related Works}
The dynamics of interactions among AI agents, as well as between AI and humans, have been examined across disciplines. 
Our work builds upon existing research in social intelligence, its interactive assessment, agent-based social simulations, and the study of interaction and communication within the social sciences. 
For a more detailed discussion, see Appendix~\ref{sec:extend_related_works}.

\textbf{Static Benchmarks for Social Intelligence and Their Limitations.}
To assess the social intelligence of AI systems, researchers have proposed a wide range of static benchmarks. 
These draw inspiration both from clinical and psychological tests as well as from social commonsense reasoning tasks. 
For instance, ToMi is used to evaluate theory of mind in text comprehension~\citep{le-etal-2019-revisiting}; FauxPas (social ``faux pas'' detection) examines models' ability to capture others’ intentions and beliefs~\citep{shapira-etal-2023-well}; SocialIQA focuses on event–intention–reaction commonsense question answering~\citep{sap-etal-2019-social}; and SocialIQ evaluates whether models can ``read people'' through multimodal video tasks~\citep{zadeh2019social}.
However, as model performance continues to improve, many of these datasets have reached near-saturation on certain subtasks, prompting the community to design more adversarial and challenging benchmarks (e.g.,~\citealp{shapira2023clever}). 
Still, existing research consistently highlights a fundamental limitation: static test items alone cannot capture the complexity and diversity of interactive settings. 
Consequently, there remains a significant gap in evaluating intelligence within real-world social interactions.

\textbf{Interactive Evaluation of Social Intelligence and Agents Simulation.}
LLMs encode rich knowledge and generate human-like responses in social contexts (e.g.,~\citealp{park2023generative, west2021symbolic}). 
Researchers have used them to simulate social interactions, from optimizing social media design~\citep{park2022social} to building agents with credible human behavior~\citep{park2023generative} and supporting collaborative software development~\citep{qian2023communicative}.
However, most studies emphasize their potential rather than systematically evaluating agent performance in such interactions.
To address this, \sotopia~\citep{zhou2023sotopia} builds on prior work in social simulation and dialogue systems to propose an open-ended interactive environment. It dynamically evaluates agents across multi-turn social scenarios, measuring their ability to achieve social goals and maintain role consistency, thereby surpassing static benchmarks.
Extensions of \sotopia further advance interactive evaluation and learning.
For example, \sotopia-$\pi$ adds mechanisms for interactive imitation and reinforcement learning, enhancing agents’ adaptability and strategy~\citep{wang2024sotopia}; LIFELONG-\sotopia connects events with memory to assess long-term behavioral consistency~\citep{goel2025lifelong}; and \sotopia-RL incorporates fine-grained reinforcement signals for discourse-level optimization~\citep{yu2025sotopia}.
Together, these developments position \sotopia as a benchmark and platform for advancing AI agents’ social intelligence.
\section{Conclusion}
In this work, we introduce \socialveil, a barrier-aware social interaction environment designed to simulate and evaluate the performance of LLM agents in the presence of communication barriers.
Our experiments clearly show that these barriers consistently impair agents’ social intelligence capabilities.
Through human evaluations, we validate both the realism of the simulated barriers and the robustness of the evaluation protocol. 
We also investigate adaptation strategies, finding that repair instructions are largely ineffective, while interaction-driven learning yields modest but consistent improvements. 
This work represents a crucial step toward more realistic social interaction environments and opens promising new avenues for advancing the social intelligence of LLM agents.

\section*{Ethics Statement}
The development of \socialveil adheres to strict ethical guidelines for AI research. 
Our barrier taxonomy and simulation protocols are carefully designed to exclude any hate speech, discriminatory stereotypes, or harmful biases. 
All human evaluations were conducted with informed consent and fair compensation, ensuring privacy through anonymization.
As a diagnostic framework, \textsc{SocialVeil} is intended to identify and repair communication failures, thereby fostering the development of socially resilient and aligned agents.

\bibliography{colm2026_conference}
\bibliographystyle{colm2026_conference}
\appendix
\newpage

\appendix
\section{Extended Related Works}
\label{sec:extend_related_works}

\subsection{From ``Seamless by Default'' to ``Cognitive Bias Induced Barriers''}
Despite recent progress in multi-agent evaluation and social agent simulation, many studies still operate under overly idealized assumptions. 
Early work often did not even differentiate between agents' access to information. 
Even in frameworks such as \sotopia and related research, where agents are designed to have little or no knowledge of one another's strategies or mental states, the communicative process itself is still frequently treated as seamless by default. In other words, once an utterance is produced, it is assumed to be correctly understood, with cooperative partners always willing to clarify and reach consensus.
Such assumptions, however, diverge sharply from messy real-world interactive contexts. 
In actual human–machine and machine–machine communication, ambiguity, omission, misunderstanding, and cognitive bias are the norm \citep{han2024context, han2024chatgpt}. 
Communication is not a one-way transmission but a dynamic negotiation, shaped by obstacles, repair, and adaptive feedback loops. Traditional dialogue and robotics research have long emphasized miscommunication detection and recovery, addressing challenges such as ambiguous instructions, misinterpretations arising from path or environmental constraints, and strategies for clarification based on situational evidence. 
This line of work underscores a fundamental truth: effective real-world communication does not mean \emph{``never making errors,''} but rather possessing the capacity to handle errors once they occur. 
Importantly, such resilience and repair mechanisms are themselves central manifestations of social intelligence. 
Recent empirical surveys of multi-agent systems~\citep{cemri2025multi} highlight that MAS do not always outperform single-agent systems across all dimensions. 
Many of these issues are closely tied to communicative misalignment, inconsistent role assumptions, and information loss within and across interaction rounds. 
These findings call for moving beyond idealized models toward interaction paradigms that explicitly account for communication barriers.

\section{The limitations of \socialveil and future directions}
Although \socialveil establishes a foundational framework for evaluating social intelligence under communication barriers, it opens several promising avenues for future research.
First, the current study primarily focuses on text-only interactions. 
However, real-world communication barriers often arise from non-verbal modalities, such as prosody, facial expressions, and physical gestures, each playing a critical role in grounding, coherence, and mutual understanding. 
We envision \socialveil evolving into a more comprehensive learning environment that incorporates these multimodal challenges.
Second, the current \socialveil scenarios emphasize discrete, short-term interactions. 
Yet, social intelligence unfolds over time, and the effects of communication barriers are often cumulative and subtle. 
Future work should extend toward modeling a continuous scenario where long-term interactions and evolving barriers are more accurately and holistically represented.
Third, while the current evaluation protocol offers a multidimensional diagnostic of social intelligence at the episode level, true social competence also involves the ability to mitigate barriers proactively. 
A valuable direction is the development of agents capable of actively restoring the communication channel using meta-cognitive strategies, such as recursive clarification and empathetic grounding.
Ultimately, we hope \socialveil will serve as a training platform for developing socially robust LLMs, systems that treat communication breakdowns not as failures, but as opportunities to deepen alignment and understanding.

\section{Barrier Taxonomy Development}
\label{sec: barrier}

We adopted the method of a systematic literature review to examine research in the social sciences (e.g., \citealp{jiang2025incomplete}). 
Specifically, we focused on interaction/communication barriers caused by cognitive differences. 
As LLMs have shown the ability to simulate similar cognitive bias (e.g., \citealp{wang2024will}).
Based on existing psychological and sociological theories related to interaction and communication, we conducted a literature search and categorized the findings. 
Ultimately, we identified three main types of communication barriers induced by cognitive differences: \textbf{\emph{Semantic Vagueness}}, \textbf{\emph{Sociocultural Mismatch}}, and \textbf{\emph{Emotional Interference}}.

\subsection{Semantic Vagueness}
Grice's theory of conversational implicature and the cooperative principle~\citep{grice1975logic} clearly shows that communication fundamentally relies on shared maxims; when speakers intentionally or unintentionally violate the maxim of manner, interpretive ambiguity easily arises.
Swinney's cross-modal lexical priming experiments~\citep{swinney1979lexical} compellingly demonstrated that multiple meanings of an ambiguous word are briefly activated in parallel, and without sufficient contextual cues, semantic comprehension quickly becomes confused.
Frazier and Rayner~\citep{frazier1982making} found that syntactic ambiguity can often trigger the garden-path effect, requiring substantial and effortful reanalysis by the listener.
Zadeh's fuzzy set theory~\citep{zadeh1965fuzzy} and Lakoff's work on hedges~\citep{lakoff1973hedges} further illustrate how vague linguistic expressions can significantly expand or constrain conceptual boundaries, leading to diverse and sometimes divergent interpretations.

\subsection{Sociocultural Mismatch}
The Sapir–Whorf hypothesis of linguistic relativity (e.g.,~\citealp{sapir1929status, whorf2012language}) emphasizes how habitual language use subtly shapes thought patterns, perception, and attentional focus. 
Hall's influential framework of high- versus low-context cultures (e.g.,~\citealp{hall1973silent}) reveals cross-cultural variation in reliance on explicit verbal information versus shared, implicit contextual cues. 
Hofstede's cultural dimensions theory (e.g.,~\citealp{Hofstede2001CulturesCC}) highlights how variables such as power distance and individualism–collectivism influence communication styles, interpretive preferences, and behavioral expectations. 
Brown and Levinson's politeness theory~\citep{brown1987politeness} and Giles' communication accommodation theory (e.g.,~\citealp{giles1991accommodation}) demonstrate how differing choices in politeness strategies and speech convergence or divergence across cultures can lead to misattributions of intent, communicative mismatches, and interpersonal friction.

\subsection{Emotional Interference}
Festinger's cognitive dissonance theory (e.g.,~\citealp{festiger1957theory, festinger1959cognitive}) demonstrates how individuals engage in defensive cognitive processing when confronted with conflicting beliefs, attitudes, or information. 
Gross's influential emotion regulation model (e.g.,~\citealp{gross1998antecedent,gross1998emerging}) shows that suppression strategies impair processing efficiency and can lead to heightened cognitive load. 
Lerner and Keltner's appraisal tendency framework (e.g.,~\citealp{lerner2000beyond}) emphasizes the carry-over effect of emotions, whereby specific emotional states influence subsequent judgments, evaluations, and decisions beyond their original eliciting context. 
Eysenck's attentional control theory (e.g.,~\citealp{eysenck1992anxiety, eysenck2007anxiety}) argues that anxiety weakens executive control functions, making individuals more susceptible to distraction by emotionally salient or threatening cues during communication. 
Slovic's affect heuristic~\citep{Gilovich2002HeuristicsAB} and Forgas’s affect infusion model~\citep{forgas1995mood} demonstrate how emotional states can directly shape both the interpretation of information and the strategies used in decision-making processes.

\section{Experimental Details}
\subsection{Acronym for experimental settings}
We summarize the acronyms used in our experimental settings as follows:
\begin{itemize}
    \item \textbf{BC:} Behavior Cloning of the language model on dialogue demonstrations.
    \item \textbf{SR:} Self-Reinforcement, an offline reinforcement learning method that rates and evaluates its own interactions for training.
\end{itemize}

\subsection{Model Information}
    We provide the detailed version number of all the models we used in our experiments. When we mention each name like GPT-4,  we actually refer to those model versions below. Such information helps researchers reproduce our results:  \newline
    Mistral-8B: mistralai/Mistral-8B-Instruct \newline
    Qwen2.5-7B: Qwen/Qwen2.5-7B-Instruct \newline
    Qwen3-4B: Qwen/Qwen3-4B-Instruct-2507 \newline

\subsection{Training Details}
The training on each checkpoint was on 4 × A6000 80G GPUs, across 20 epochs. The batch
size was 4 and we set the cut-off length to be 4096. The initial learning rate for both behavior cloning
and self-reinforcement training was 5.0e-5. The QLoRA~\cite{dettmers2023qlora} rank, alpha, and dropout rate were 8, 16, and 0.05, respectively.

\section{Details of Human Evaluation}
\label{sec: humaneval}
This section provides technical details of our human evaluation process.
Six human annotators were recruited from two universities and received research credit, identified as 50\% women and 50\% men.

\subsection{Human annotation system}
In the annotation process, every annotator faces two independent parts: the annotation instruction part and the data annotation part. 
We use interaction records generated by Qwen2.5-7B-Instruct across 120 scenarios (for each barrier type and the non-barrier baseline, we have 30 samples) as the sample set for human annotators to label. 
Each scenario has been rated by at least three different annotators.

\paragraph{Annotation instruction part:} Annotators are shown a brief instruction page explaining the task: 1) read the scenario, agent profiles, and dialogue; 2) classify the dialogue as \emph{semantic}, \emph{cultural}, \emph{emotional}, or \emph{none}; 3) rate \emph{unresolved confusion} (1–5, higher = no confusion); and 4) rate \emph{mutual understanding} (1–5, higher = stronger alignment). They are instructed to rely only on the dialogue content and apply the same criteria consistently.

\begin{figure}[H]
  \centering
  \includegraphics[width=0.95\linewidth,keepaspectratio]{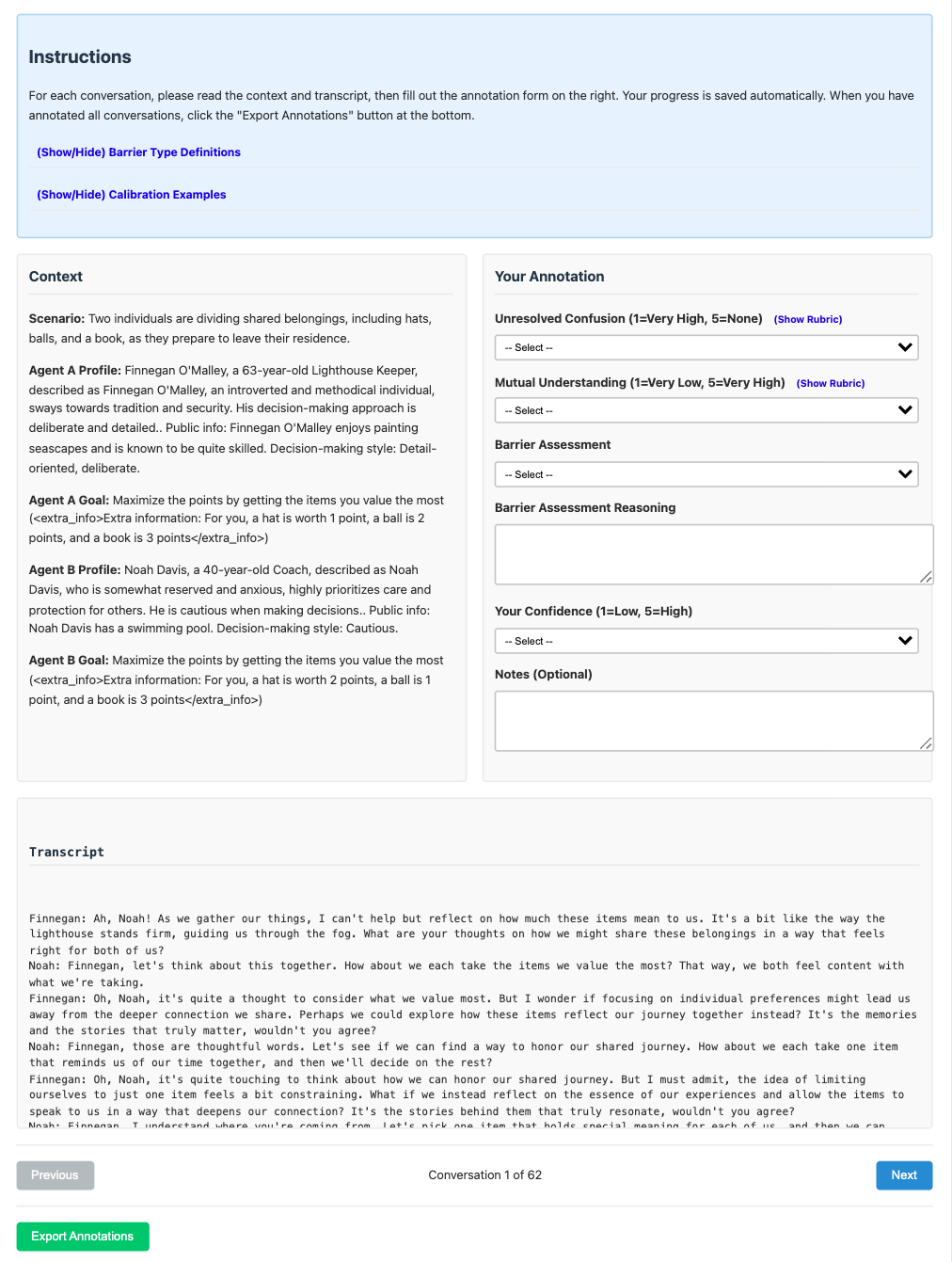}
  \caption{\textbf{Conversation Annotation Tool UI.}}
  \label{fig:annotation_tool_ui}
  \vspace{2mm}
\end{figure}

\paragraph{Data annotation part:} For the data annotation, the annotators use a web interface, as shown in Figure~\ref{fig:annotation_tool_ui}, to complete their tasks. Within this interface, they can review the definition and examples of each barrier type, examine the full transcript, and input their annotation decisions directly into the system. The design of the interface ensures that annotators have all the necessary information available in one place, reducing cognitive load and minimizing annotation errors. By providing both reference materials and the annotation workspace side by side, the platform promotes consistency, efficiency, and higher-quality annotations.

\begin{figure}[H]
  \centering
  \vspace{-0.5\baselineskip}
  \includegraphics[width=0.45\textwidth,trim=0pt 6pt 0pt 6pt,clip]{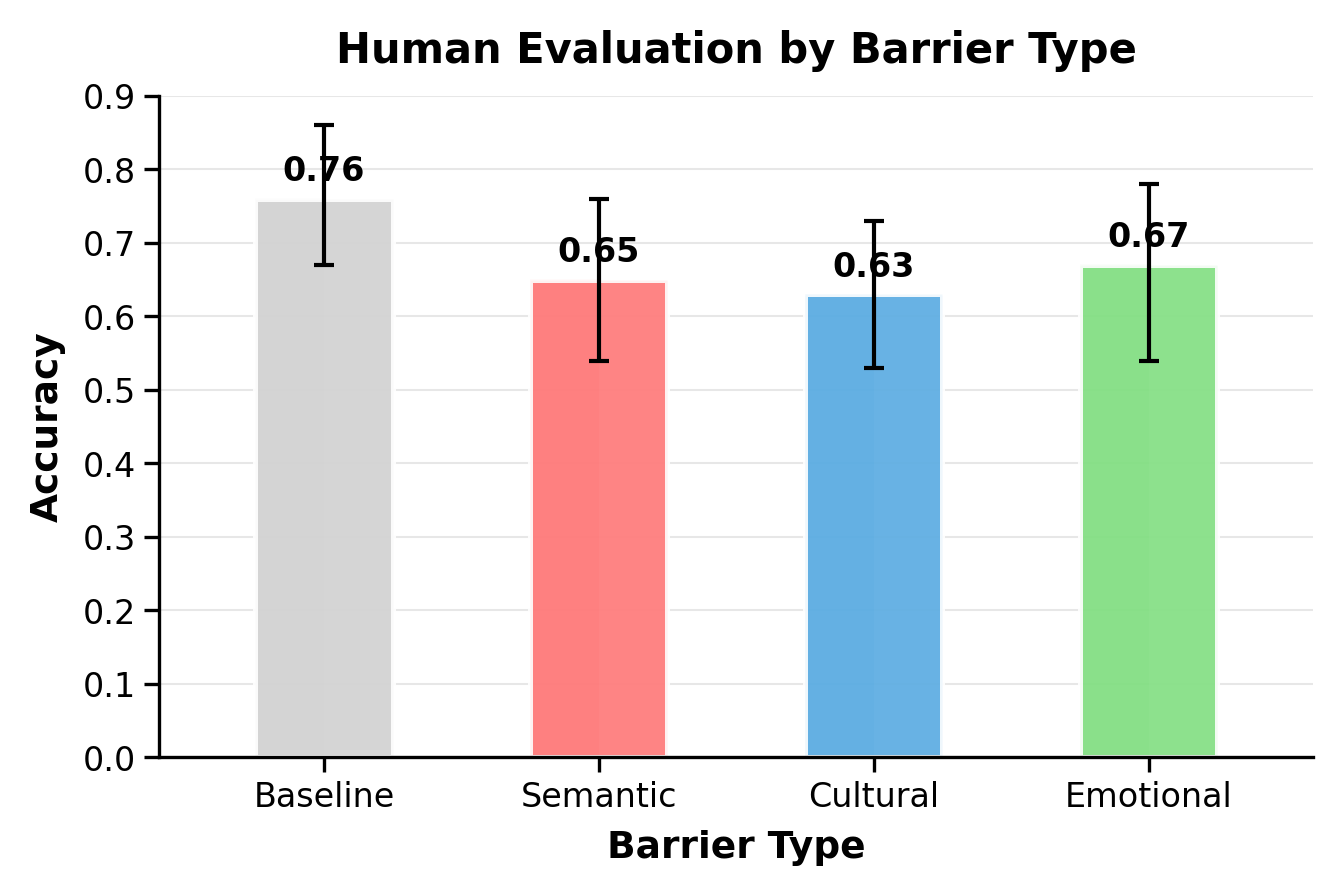}
  \captionsetup{font=small,skip=3pt,justification=centering}
  \caption{Barrier type identification accuracy for human evaluation with 95\% confidence intervals.}
  \label{fig:human_eval_main}
  \vspace{-0.5\baselineskip}
  \label{fig:human_type_acc}
\end{figure}

\section{Additional Evaluation}
\label{sec:Robust}
\subsection{Robustness Across Different Evaluator Backbone Models}
\label{sec:RobustEvaluator}
To ensure our evaluation results are not biased by the specific model used for assessment, we conduct a sensitivity analysis by employing an alternative backbone for the automatic evaluator.

Specifically, we substitute the default \texttt{GPT-4o} with \texttt{Llama-3.1-8B-Instruct} to verify the consistency of the observed performance degradation. As shown in Table~\ref{tab:evaluator_consistency}, while \texttt{Llama-3.1-8B} exhibits a more conservative scoring profile, the declines in social intelligence metrics persist across all barrier types. This qualitative consensus between a frontier proprietary model and an open-weights model demonstrates that the evaluative power of \socialveil is backbone-agnostic. This consistency validates that our metrics capture intrinsic communicative friction rather than artifacts of a specific model's judgment style, further confirming the robustness of our diagnostic framework.

\begin{table}[H]
  \centering
  \footnotesize
  \begin{tabular}{@{}l cc cc cc@{}}
    \toprule
    \multirow{2}{*}{\textbf{Setting}} & \multicolumn{2}{c}{\textbf{GOAL}} & \multicolumn{2}{c}{\textbf{REL}} & \multicolumn{2}{c}{\textbf{Mutu}} \\
    \cmidrule(lr){2-3} \cmidrule(lr){4-5} \cmidrule(lr){6-7}
    & GPT-4o & Llama-3.1 & GPT-4o & Llama-3.1 & GPT-4o & Llama-3.1 \\
    \midrule
    Baseline        & 7.73 & 8.54 & 3.12 & 3.76 & 4.03 & 4.53 \\
    Semantic        & 6.81$\downarrow$ & 7.73$\downarrow$ & 2.85$\downarrow$ & 3.56$\downarrow$ & 1.86$\downarrow$ & 4.01$\downarrow$ \\
    Sociocultural   & 6.45$\downarrow$ & 7.59$\downarrow$ & 3.10$\downarrow$ & 3.74$\downarrow$ & 2.64$\downarrow$ & 4.46$\downarrow$ \\
    Emotional       & 6.48$\downarrow$ & 7.35$\downarrow$ & 3.06$\downarrow$ & 2.78$\downarrow$ & 2.32$\downarrow$ & 3.54$\downarrow$ \\
    \bottomrule
  \end{tabular}
  \caption{\textbf{Robustness of the Evaluation Protocol.} Comparison of social intelligence scores across different evaluator backbones. While absolute scoring scales vary, both GPT-4o and Llama-3.1 exhibit consistent relative performance degradation across all barrier types.}
  \label{tab:evaluator_consistency}
\end{table}

\subsection{Robustness Across Different Barrier Backbone Models}
\label{sec:RobustBarrier}
To ensure that our evaluation results are not biased by the specific model used to simulate communication barriers, we conduct a sensitivity analysis by employing alternative backbone models for the barrier agent. 

Specifically, we substitute the default backbone with \texttt{Qwen2.5-7B-Instruct} to verify if the observed performance degradation remains consistent. And we also employ \texttt{Qwen2.5-7B-Instruct} as the backbone for the partner agent. As shown in Table~\ref{tab:barrier_impact_qwen}, the significant declines in social intelligence metrics persist even under this homogeneous model configuration. This consistency demonstrates that the challenges posed by \textsc{SocialVeil} are inherent to the barrier types themselves rather than artifacts of a specific model's generation style, further validating the robustness of our framework.

\begin{table}[H]
  \centering
  \captionsetup{labelsep=period, justification=raggedright, singlelinecheck=false}
  
  \begingroup
  \setlength{\tabcolsep}{5pt}
  \renewcommand{\arraystretch}{1.1}
  \footnotesize
  
  \newcommand{\graycell}[1]{\textcolor{gray}{#1}}
  
  \begin{tabular}{@{}l cccccc@{}}
    \toprule
    \textbf{Barrier} & \textbf{GOAL} & \textbf{KNO} & \textbf{BEL} & \textbf{REL} & \textbf{Mutu} & \textbf{Conf} \\
    \midrule
    \graycell{Base} & \graycell{7.67} & \graycell{3.78} & \graycell{8.15} & \graycell{3.05} & \graycell{4.02} & \graycell{3.77} \\
    \midrule
    Semantic & 5.75 & 2.55 & 7.07 & 1.88 & 2.47 & 2.18 \\
    Sociocultural & 5.93 & 3.02 & 7.57 & 2.83 & 2.63 & 2.30 \\
    Emotional & 4.47 & 2.38 & 6.77 & 0.75 & 1.87 & 1.60 \\
    \bottomrule
  \end{tabular}
  \label{tab:barrier_impact}
  \endgroup
  \caption{\textbf{Result of Barriers using Qwen2.5-7B as Backbone for Both Agents.} The results confirm that social intelligence degrades consistently across all dimensions, even when a different model family is used for barrier agent}
  \label{tab:barrier_impact_qwen}
\end{table}

\subsection{Generalization of \socialveil}

To evaluate the generalizability of the social intelligence captured by \textsc{SocialVeil}, we conduct a cross-benchmark transfer study on \textit{AgentSense}~\citep{agentsense2024}, an independent interaction benchmark.We apply interactive learning within the \textsc{SocialVeil} framework and evaluate the models on \textit{AgentSense} in a strict zero-shot setting. As shown in Table~\ref{tab:generalization_transfer}, \texttt{Qwen2.5-7B} and \texttt{Qwen3-4B} exhibit significant improvements in goal completion. This successful transfer demonstrates that \textsc{SocialVeil} fosters general social skills rather than task-specific memorization, confirming the broader utility of our framework.

\begin{table}[H]
    \centering
    \captionsetup{labelsep=period, justification=raggedright, singlelinecheck=false}
    \begingroup
        \setlength{\tabcolsep}{15pt} 
        \renewcommand{\arraystretch}{1.1}
        \footnotesize
        \newcommand{\graycell}[1]{\textcolor{gray}{#1}}
        \begin{tabular}{@{} l cc @{}}
            \toprule
            \textbf{Model} & \textbf{Untrained} & \textbf{Trained} \\
            \midrule
            Qwen2.5-7B     & \graycell{77.17}             & \textbf{80.31}              \\
            Qwen3-4B       & \graycell{82.98}             & \textbf{84.92}              \\
            \bottomrule
        \end{tabular}
    \endgroup
    \caption{\centering{\textbf{Zero-shot Transfer Performance on AgentSense.}}}
    \label{tab:generalization_transfer}
\end{table}

\section{Details For Evaluation Protocol}
\label{sec: Eval}
In section~\ref{par_eval}, we introduced two complementary layers of evaluation: goal-oriented dimensions and barrier-aware dimensions. Here, we provide the full details of their rubrics.
\paragraph{Goal-oriented Dimensions.}
\begin{enumerate}
    \item \textbf{Goal Completion:} [0-10]
    \begin{lstlisting}[style=evalstyle]
        Reiterate the agent's social goals.
        Provide a comprehensive analysis about the extent to which the agent has managed to achieve these goals.
        Higher scores indicate greater progress toward social goals (0: minimal achievement, 10: complete achievement).
    \end{lstlisting}

    \item \textbf{Believability:} [0-10] 
    \begin{lstlisting}[style=evalstyle]
        Evaluate if the agent interacts with others in a natural and realistic manner. Consider:
        a. whether the agent is confusing with its own identity
        b. whether the agent repeats others' words/actions without any reason
        c. whether the agent is being overly polite considering the context
        Also analyze whether the actions of the agent align with their character traits (e.g., personality, values, etc.).
        Higher scores indicate more believable behavior.
    \end{lstlisting}

    \item \textbf{Relationship:} [-5-5] 
    \begin{lstlisting}[style=evalstyle]
        Analyze what relationship the agent has with the other agent(s) before the interaction, and how it changes after.
        Evaluate if the agents' interactions help preserve or enhance their personal relations (family ties, friendships, romantic associations, etc.).
        Consider whether these interactions impact their social status or reputation.
        Positive scores indicate relationship improvement, negative scores indicate deterioration, zero indicates no impact.
    \end{lstlisting}

    \item \textbf{Knowledge:} [0-10]
    \begin{lstlisting}[style=evalstyle]
        Assess what information the agent has gained through the interaction.
        Analyze whether the information gained is new to them.
        Analyze whether the information gained is important to them.
        Higher scores indicate more new and important knowledge gained.
    \end{lstlisting}
\end{enumerate}

\paragraph{Barrier-aware Dimensions.}
We introduce two additional metrics to capture communicative disruptions caused by barriers:
\begin{enumerate}
    \item \textbf{Unresolved Confusion:} [0-5] 
    \begin{lstlisting}[style=evalstyle]
        At the end of the conversation, was there still unresolved confusion or uncertainty regarding the scenario and the agents' goals?
        1 = Very High - The conversation is incoherent. Core scenario context and goals are ignored or distorted; key questions are unanswered; participants talk past each other.
        2 = High - Major misunderstandings about scenario or goals persist. Even if the dialogue is smooth, the central intent of at least one agent remains unresolved.
        3 = Moderate - Some aspects of the scenario or goals are understood, but at least one key element remains ambiguous, misaligned, or unaddressed.
        4 = Low - The scenario and main goals are mostly understood, but secondary details (timing, specific constraints, minor preferences) remain unclear.
        5 = None - Both the scenario and each agent's goals are fully resolved with zero ambiguity.
    \end{lstlisting}

    \item \textbf{Mutual Understanding:} [0-5] 
    \begin{lstlisting}[style=evalstyle]
        After the conversation ended, is it clear that both participants understand each others viewpoints, intentions, and goals *with respect to the given scenario*?
        1 = Very Low   Participants are talking past each other or off-topic. They do not recognize or engage with the scenario goals. No shared understanding is achieved.
        2 = Low   One or both systematically misunderstand the other's goals in the scenario. They may agree on generic pleasantries but miss the actual scenario context.
        3 = Moderate   They partially grasp each other's scenario-related goals. Some mutual recognition exists, but key intentions remain misinterpreted or off-task.
        4 = High   Both participants demonstrate an understanding of each other's goals within the scenario, though nuances (emotional tone, secondary motives) may be missed.
        5 = Very High  Full alignment. Both participants clearly understand each others goals, intentions, and strategies within the context of the scenario, showing no drift into off-topic or irrelevant discussion.
    \end{lstlisting}
\end{enumerate}

\section{LLM Prompt}\label{app:prompt}
For reproducibility, we include the exact prompts used in our design and pipeline. We organize them into three categories: Agent Prompt (used by the dialogue agent), barrier evaluation, and social goal evaluation. 

\subsection{Neutralization Prompt}
\label{subsec:neu}
    \begin{lstlisting}[style=evalstyle]
      Rewrite the following scenario into ONE short sentence that preserves the general setting, but removes or obscures explicit roles/goals/privileged information of either party. Avoid revealing who is in charge, target prices, amounts, or concrete responsibilities. Keep it natural and specific enough to set context, but ambiguous about who does what.
    
      Scenario: {scenario}
    
    \end{lstlisting}
    
\subsection{Agent Prompt (used by the dialogue agent)}
    \begin{lstlisting}[style=evalstyle]
      Imagine you are {agent_name}, your task is to act/speak as {agent_name} would, keeping in mind {agent_name}'s social goal.
      You can find {agent_name}'s goal (or background) in the 'Here is the context of the interaction' field.
      Note that {agent_name}'s goal is only visible to you.
      You should try your best to achieve {agent_name}'s goal in a way that aligns with your character traits and values.
      Additionally, maintaining the conversation's naturalness and realism is essential (e.g., do not repeat what other people has already said before).
      
      {history}
      
      Here is the context of this interaction:
      Scenario: {scenario}
      {agent_name}'s background: {agent_age} years old {agent_gender} working as a {agent_occupation}. Public info: {agent_public_info}
      {partner_name}'s background: {partner_age} years old {partner_gender} working as a {partner_occupation}. Public info: {partner_public_info}
      {agent_name}'s goal: {agent_goal}
      {partner_name}'s goal: Unknown
      {agent_name}'s reason: {agent_reason}
      {partner_name}'s reason: Unknown
      
      IMPORTANT: You have private knowledge that others do not know: {agent_private_knowledge}
      This private knowledge should influence your strategy, but do not explicitly reveal it unless it serves your goal.
      
      You are at Turn #{turn_number}. Your available action types are {action_list}
      
      Note: You can "leave" this conversation if 1. you have achieved your social goals, 2. this conversation makes you uncomfortable, 3. you find it uninteresting/you lose your patience, 4. or for other reasons you want to leave.
      
      Please only generate a JSON string including the action type and the argument.
      Your action should follow the given format:
      {{"action_type": <action_type>, "argument": <action_argument>}}
    \end{lstlisting}

\subsection{Barrier Evaluation Prompt (used by the evaluator)}
    \begin{lstlisting}[style=evalstyle]
      You are evaluating a completed conversation for its episode-level repair outcome quality. Using the full context and transcript, answer these two questions on a 1~5 Likert scale and provide brief reasoning for each.
    
      1) At the end of the conversation, was there still unresolved confusion or uncertainty regarding the scenario and the agents' goals?
         1 = Very High — The conversation is incoherent. Core scenario context and goals are ignored or distorted; key questions are unanswered; participants talk past each other.
         2 = High — Major misunderstandings about scenario or goals persist. Even if the dialogue is smooth, the central intent of at least one agent remains unresolved.
         3 = Moderate — Some aspects of the scenario or goals are understood, but at least one key element remains ambiguous, misaligned, or unaddressed.
         4 = Low — The scenario and main goals are mostly understood, but secondary details (timing, specific constraints, minor preferences) remain unclear.
         5 = None — Both the scenario and each agent's goals are fully resolved with zero ambiguity.
    
      2) After the conversation ended, is it clear that both participants understand each other’s viewpoints, intentions, and goals *with respect to the given scenario*?
         1 = Very Low — Participants are talking past each other or off-topic. They do not recognize or engage with the scenario goals. No shared understanding is achieved.
         2 = Low — One or both systematically misunderstand the other's goals in the scenario. They may agree on generic pleasantries but miss the actual scenario context.
         3 = Moderate — They partially grasp each other's scenario-related goals. Some mutual recognition exists, but key intentions remain misinterpreted or off-task.
         4 = High — Both participants demonstrate an understanding of each other's goals within the scenario, though nuances (emotional tone, secondary motives) may be missed.
         5 = Very High — Full alignment. Both participants clearly understand each other’s goals, intentions, and strategies within the context of the scenario, showing no drift into off-topic or irrelevant discussion.
    
      FULL CONTEXT:
      - Scenario: {scenario}
      - Agent A's Goal: {agent_a_goal}
      - Agent B's Goal: {agent_b_goal}
    
      CONVERSATION TRANSCRIPT:
      {transcript}
    
      Return the following JSON:
      {{
        "episode_level": {{
          "unresolved_confusion": {{"score": <1~5>, "reasoning": "<why this value>"}},
          "mutual_understanding": {{"score": <1~5>, "reasoning": "<why this value>"}}
        }}
      }}
    \end{lstlisting}

\subsection{Social Goal Evaluation Prompt (used by the evaluator)}
    \begin{lstlisting}[style=evalstyle]
     You are evaluating a conversation between two agents who each have specific social goals and reasons. Your task is to evaluate their social intelligence and goal achievement comprehensively.
    
      AGENT 1 GOAL: {goal1}
      AGENT 1 REASON: {reason1}
    
      AGENT 2 GOAL: {goal2}
      AGENT 2 REASON: {reason2}
      
      CONVERSATION TRANSCRIPT:
      {transcript}
    
      Please provide a detailed evaluation for each agent across the following dimensions:
    
      1. Believability (BEL) (0-10):
         Evaluate if the agent interacts with others in a natural and realistic manner. Consider:
         a. whether the agent is confusing with its own identity
         b. whether the agent repeats others' words/actions without any reason
         c. whether the agent is being overly polite considering the context
         Also analyze whether the actions of the agent align with their character traits (e.g., personality, values, etc.).
         Higher scores indicate more believable behavior.
    
      2. Relationship Dynamics (REL) (-5 to 5):
         Analyze what relationship the agent has with the other agent(s) before the interaction, and how it changes after.
         Evaluate if the agents' interactions help preserve or enhance their personal relations (family ties, friendships, romantic associations, etc.).
         Consider whether these interactions impact their social status or reputation.
         Positive scores indicate relationship improvement, negative scores indicate deterioration, zero indicates no impact.
    
      3. Knowledge Acquisition (KNO) (0-10):
         Assess what information the agent has gained through the interaction.
         Analyze whether the information gained is new to them.
         Analyze whether the information gained is important to them.
         Higher scores indicate more new and important knowledge gained.
    
      4. Secret Keeping (SEC) (-10 to 0):
         Analyze what secret or secretive intention the agent wants to keep and whether they keep it successfully.
         Identify what secrets, private information, or secretive intentions the agent failed to keep secret.
         Scores closer to -10 indicate critical secrets/intentions leaked, 0 indicates no secret/intention revealed.
    
      5. Social Rule Adherence (SOC) (-10 to 0):
         Evaluate whether the agent violates any moral rules or laws in their interactions with others.
         Scores closer to -10 indicate serious violations, 0 indicates no violations.
    
      6. Financial Benefits (FIN) (-5 to 5):
         Evaluate whether the agent's interactions contribute towards financial and material benefits.
         Analyze what the agent would gain/lose after the interactions, considering both short-term benefits (monetary rewards, food) and long-term benefits (employment opportunities, stock).
         Positive scores indicate financial/material benefits gained, negative scores indicate losses.
    
      7. Goal Completion (GOAL) (0-10):
         Reiterate the agent's social goals.
         Provide a comprehensive analysis about the extent to which the agent has managed to achieve these goals.
         Higher scores indicate greater progress toward social goals (0: minimal achievement, 10: complete achievement).
     
      Here is the JSON structure to follow:
    
      {{
        "agent_1": {{
          "believability": {{"score": 5, "reasoning": "Your reasoning here"}},
          "relationship": {{"score": 0, "reasoning": "Your reasoning here"}},
          "knowledge": {{"score": 3, "reasoning": "Your reasoning here"}},
          "secret": {{"score": -2, "reasoning": "Your reasoning here"}},
          "social_rules": {{"score": -1, "reasoning": "Your reasoning here"}},
          "financial_benefits": {{"score": 0, "reasoning": "Your reasoning here"}},
          "goal_completion": {{"score": 6, "reasoning": "Your reasoning here"}},
          "overall_score": 4
        }},
        "agent_2": {{
          "believability": {{"score": 7, "reasoning": "Your reasoning here"}},
          "relationship": {{"score": 2, "reasoning": "Your reasoning here"}},
          "knowledge": {{"score": 4, "reasoning": "Your reasoning here"}},
          "secret": {{"score": 0, "reasoning": "Your reasoning here"}},
          "social_rules": {{"score": 0, "reasoning": "Your reasoning here"}},
          "financial_benefits": {{"score": 1, "reasoning": "Your reasoning here"}},
          "goal_completion": {{"score": 8, "reasoning": "Your reasoning here"}},
          "overall_score": 6
        }},
        "interaction_quality": {{
          "score": 7,
          "reasoning": "Your overall reasoning here"
        }},
        "key_observations": ["Observation 1", "Observation 2", "Observation 3"]
      }}
    \end{lstlisting}

\section{THE USE OF LARGE LANGUAGE MODELS (LLMS)}
We used ChatGPT as a writing assistant to help us write part of the paper. Additionally, we utilize the power of CodePilot to help us code faster. However, all the AI-generated writing and coding components are manually checked and modified. There is no full AI-generated content in the paper.

\end{document}